\documentclass[letterpaper]{article} 
\usepackage{aaai2026}  
\usepackage{times}  
\usepackage{helvet}  
\usepackage{courier}  
\usepackage[hyphens]{url}  
\usepackage{graphicx} 
\usepackage{natbib}  
\usepackage{caption} 
\usepackage{algorithm}
\usepackage{algorithmic}
\usepackage{booktabs}
\usepackage{multirow}
\usepackage{lipsum}
\usepackage{newfloat}
\usepackage{listings}
\DeclareCaptionStyle{ruled}{labelfont=normalfont,labelsep=colon,strut=off} 
\floatstyle{ruled}
\newfloat{listing}{tb}{lst}{}
\floatname{listing}{Listing}
\usepackage{amsmath}
\usepackage{amssymb}
\usepackage{dsfont}
\usepackage{amsthm}
\newtheorem{definition}{DEFINITION}
\usepackage{mathrsfs}
\usepackage{subcaption}
\usepackage{enumitem}
\usepackage{xcolor}
\usepackage{booktabs} 
\usepackage{multirow} 
\usepackage{graphicx}
\newtheorem{lemma}{Lemma}

\nocopyright 

\title{GeoPTH: A Lightweight Approach to Category-Based Trajectory Retrieval via Geometric Prototype Trajectory Hashing}
\author {
    Yang Xu\thanks{These authors contributed equally.},
    Zuliang Yang\footnotemark[1], 
    Kai Ming Ting\thanks{Corresponding author.}
}
\affiliations {
    State Key Laboratory for Novel Software Technology, Nanjing University, Nanjing, China\\
    School of Artificial Intelligence, Nanjing University, Nanjing,  China\\
    \{xuyang, yangzl\}@lamda.nju.edu.cn, tingkm@nju.edu.cn
}

\begin{document}

\maketitle

\begin{abstract}
    Trajectory similarity retrieval is an important part of spatiotemporal data mining, however, existing methods have the following limitations:
    traditional metrics are computationally expensive, while learning-based methods suffer from substantial training costs and potential instability.
    This paper addresses these problems by proposing \textbf{Geo}metric \textbf{P}rototype \textbf{T}rajectory \textbf{H}ashing (GeoPTH), a novel, lightweight, and non-learning framework for efficient category-based trajectory retrieval.
    GeoPTH constructs data-dependent hash functions by using representative trajectory prototypes, i.e., small point sets preserving geometric characteristics, as anchors. The hashing process is efficient, which involves mapping a new trajectory to its closest prototype via a robust, \textit{Hausdorff} metric.
    Extensive experiments show that GeoPTH’s retrieval accuracy is highly competitive with both traditional metrics and state-of-the-art learning methods, and it significantly outperforms binary codes generated through simple binarization of the learned embeddings. Critically, GeoPTH consistently outperforms all competitors in terms of
    efficiency. Our work demonstrates that a lightweight, prototype-centric approach offers a practical and powerful alternative, achieving an exceptional retrieval performance and computational efficiency.
\end{abstract}



\section{Introduction}

\begin{figure}[t]
    \centering
    \includegraphics[width=0.98\columnwidth]{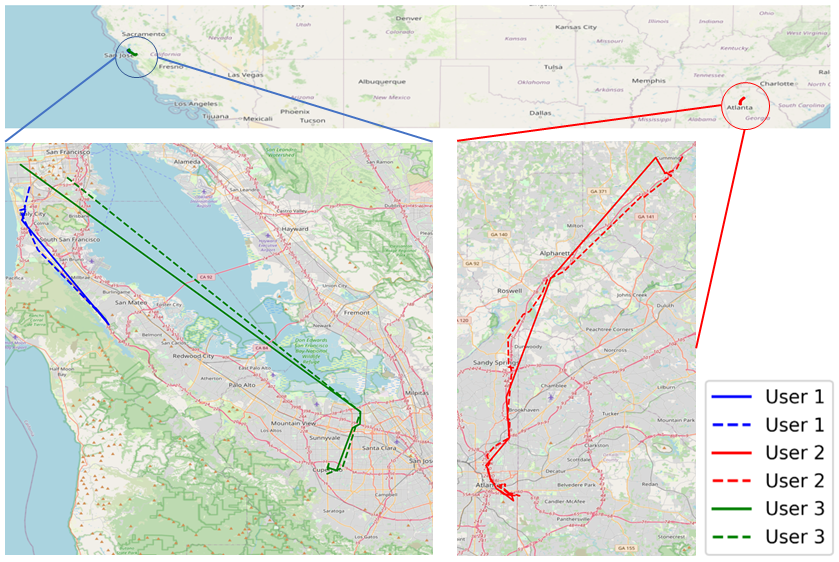}
    \caption{Behavior trajectories from three different users in the \texttt{Gowalla} dataset.}
    \label{fig:1}
\end{figure}

The proliferation of location-aware devices has led to an unprecedented explosion of trajectory data. Efficient trajectory similarity retrieval is fundamental to high-impact applications, including traffic analysis, human mobility modeling, and urban planning~\cite{hui2021trajnet,luca2021survey,ji2022precision}.
A key challenge is retrieving trajectories by their functional category or underlying behavioral pattern, a crucial step for turning unstructured data into knowledge~\cite{fang2022spatio,luo2023task}.
Measuring geometrically similar paths is crucial for this high-level, category-based trajectory retrieval, and it hinges on a general observation: trajectories with geometrically similar paths often share the same underlying behavioral patterns~\cite{jin2020trajectory,liao2024bat}.
Figure~\ref{fig:1} illustrates this observation with user behavior trajectories from the \texttt{Gowalla} dataset~\cite{Gowalla}.

The gold standard for measuring trajectory similarity has long been established by traditional, non-learning metrics such as \textit{Hausdorff distance}~\cite{hausdorff}, \textit{Fr\'echet distance}~\cite{frechet}, and \textit{Dynamic Time Warping (DTW)}~\cite{dtw}. These methods are lauded for their accuracy in capturing the geometric fidelity between two trajectories and are often treated as a benchmark for precision~\cite{hu2023spatio}. For example, \textit{Hausdorff distance} measures the dissimilarity by identifying the point on one trajectory that is farthest from any point on the other, thus capturing the maximum deviation between the two paths. However, despite their accuracy, these metrics rely on exhaustive point-wise computations, resulting in high (often quadratic) computational complexity that makes them prohibitively slow for large-scale, low-latency search~\cite{sousa2020vehicle}.

To overcome the high computational complexity of traditional metrics, learning-based measures have recently attracted substantial interest.
This approach aims to learn a mapping from a variable-length trajectory to a fixed-length vector, or embedding, thereby reducing the complex similarity computation to a simple and fast vector distance calculation. A prominent strategy, employed by methods such as \textit{GnesDA}~\cite{chang2025general}, involves training a neural network to produce embeddings whose distances approximate those calculated by a non-learned metric like \textit{Hausdorff distance}. However, although these measures are predicated on improving efficiency, a comprehensive study~\cite{chang2024trajectory} from an efficiency perspective reveals a critical shortcoming that these learning-based measures often suffer from exorbitant training costs that demand substantial time and resources. Crucially, the study finds that for online or one-off computations where embeddings cannot be pre-computed and reused, many learning-based measures are in fact significantly slower than the traditional metrics they were designed to replace. This efficiency shortcoming necessitates a new approach that can achieve high performance without the burdensome setup costs and hidden inefficiencies of the current learning-based measures.

Category hashing\footnote{Category hashing is also referred to as ``semantic hashing'' in some research. To avoid ambiguity, we uniformly use the term ``category hashing'' throughout this paper.} is a popular and highly successful technique in computer vision, where the category information of images is preserved in compact binary codes to enable efficient data compression and retrieval~\cite{wang2017survey,luo2023survey}.
As image input is typically in pixel form, with a significant gap between its low-level pixel and the high-level category information it contains, category hashing typically requires complex network structures to extract high-level features~\cite{cao2025deep}.
Inspired by this, \textit{Traj2Hash}~\cite{deng2024learning} has been explored for mapping trajectories to binary codes by using a deep network to learn embeddings that approximate traditional metrics.
However, it faces two potential issues. First, it simply uses a binarization layer on top of the learned embeddings, which can result in significant information loss. Second, like most learning-based methods, its training still requires high computational cost, limiting its utility.

Our starting point comes from a core insight: while category hashing in computer vision often requires complex network architectures to extract high-level category features, this process can be performed much more directly and efficiently for trajectory retrieval. This is because a trajectory's category information is highly correlated with its spatio-temporal geometric shape, which can be effectively captured by a traditional metric without an additional, costly learning process.
Thus, we propose \textbf{Geo}metric \textbf{P}rototype \textbf{T}rajectory \textbf{H}ashing (GeoPTH), a lightweight and non-learning framework that maps trajectories to binary codes based on geometric similarity.
GeoPTH achieves this by measuring a trajectory's proximity to a set of representative prototypes using \textit{Hausdorff distance}.
The \textit{Hausdorff distance} makes GeoPTH focus on the overall shape of the point set, and its binary metric space satisfies the triangle inequality, which can ensure the geometric fidelity in the space. Through this framework, GeoPTH achieves efficient trajectory mapping and efficient retrieval by leveraging the Hamming distance of binary codes.

Our main contributions are summarized as follows:
\begin{itemize}
    \item Proposing GeoPTH, a novel, lightweight, and non-learning framework for efficient and high-fidelity geometric similarity category hashing of trajectories.
    \item Showing that the metric space in which GeoPTH operates satisfies the triangle inequality, which provides a key guarantee that the resulting binary codes can effectively preserve geometric similarity.
    \item Demonstrating through extensive retrieval experiments on multiple real-world trajectory datasets that GeoPTH achieves performance competitive with both traditional and learning-based approaches, while substantially reducing the required computational costs.
\end{itemize}

\section{Preliminary}

In this section, we give the definition of trajectory and formalize the problem of \textit{trajectory hashing retrieval}. Table~\ref{tab:notation} shows the key notations used in this paper.

\begin{table}[t]
    \small
    \centering
    \renewcommand{\arraystretch}{1.1}
    \begin{tabular}{@{}l|l@{}}
        \toprule
        Notations                                      & Descriptions                                                                          \\
        \midrule
        $p$                                            & A point in $d$-dimensional real domain $\mathbb{R}^d$                                 \\
        $\mathcal{T}$                                  & A trajectory of $\langle p_1, \dots, p_\mu \rangle$ with $\mu = |\mathcal{T}|$ points \\
        $\mathcal{P}_\mathcal{T}$                      & Probability distribution that generates $\mathcal{T} \sim \mathcal{P}_\mathcal{T}$    \\
        $\mathcal{D}$                                  & Set of $N$ trajectories $\{\mathcal{T}_i\dots, \mathcal{T}_N\}$                       \\
        $K$                                            & A trajectory prototype                                                                \\
        $Q_m$                                          & A prototype codebook $\{K_{m,1},\dots,K_{m,\psi}\}$                                   \\
        $\mathcal{C}$                                  & The set of prototype codebooks $\{Q_1,\dots,Q_M\}$                                    \\
        $\mathbb{B}^L$                                 & $L$-bit binary encoding space                                                         \\
        $S_{\mathcal{P}}(\mathcal{T}_i,\mathcal{T}_j)$ & Category similarity between $\mathcal{T}_i$ and $\mathcal{T}_j$                       \\
        $\mathcal{H}$                                  & A hash function : $\mathcal{T} \to \mathbb{B}^L$                                      \\
        $b$                                            & $L$-bit binary code generated by $\mathcal{H}(\mathcal{T})$                           \\
        $S_\mathbb{H}, d_\mathbb{H}$                   & \textit{Hamming similarity} \& \textit{distance}                                      \\
        $d_H,d_h$                                      & \textit{Hausdorff distance} \& \textit{directed Hausdorff distance}                   \\
        \bottomrule
    \end{tabular}
    \caption{Key notations used in this paper.}
    \label{tab:notation}
\end{table}

\begin{definition}[\textbf{Trajectory}]
    A trajectory $\mathcal{T}$ is a sequence of time-ordered points in a $d$-dimensional space (typically $d=2$ for geographical coordinates), i.e., $\mathcal{T}=\langle p_1, \dots, p_i, \dots, p_\mu \rangle$, where $i\in[1,\mu]$ indicates the order of traversal in $\mathcal{T}$ and $\mu = |\mathcal{T}|$ is the length of the trajectory.
\end{definition}

Let $\mathcal{D}=\{\mathcal{T}_1,\dots,\mathcal{T}_N\}$ denote a reference trajectory database, drawn from a mixture of ${C}$ ($2\leq{C}<N$) underlying distributions $\{\mathcal{P}^1,\dots,\mathcal{P}^{C}\}$, where each distribution corresponds to a distinct trajectory category. Trajectories originating from the same distribution are considered similar, while those from different distributions are considered dissimilar.
Let $S_{\mathcal{P}}(\mathcal{T}_i,\mathcal{T}_j)\in\{0,1\}$ denote the category similarity between $\mathcal{T}_i$ and $\mathcal{T}_j$, where $S_{\mathcal{P}}(\mathcal{T}_i,\mathcal{T}_j)=1$ if $\mathcal{T}_i$ and $\mathcal{T}_j$ are drawn from the same distribution, and 0 otherwise.
The goal of trajectory hashing is to find an optimal trajectory function $\mathcal{H}^*(\cdot)$:
\begin{equation}
    \mathcal{H}^* = \underset{\mathcal{H} \in H}{\operatorname{argmin}} \, \mathbb{E}_{\mathcal{D}} \left| S_\mathbb{H}(\mathcal{H}(\mathcal{T}_i), \mathcal{H}(\mathcal{T}_j)) - S_\mathcal{P}(\mathcal{T}_i, \mathcal{T}_j) \right|,
    \label{eq:H*}
\end{equation}
where $\mathcal{H}$ is a hash function $\mathcal{H} : \mathcal{T} \to \mathbb{B}^L$ from the hash function space $H$ that maps a trajectory $\mathcal{T}$ to a $L$-bit binary code $b = \mathcal{H}(\mathcal{T})$, and $S_\mathbb{H}(\cdot,\cdot)$ is the Hamming similarity between two binary codes, defined as follows:
\begin{definition}[\textbf{Hamming similarity}]
    Given two $L$-bit binary codes $b_i, b_j \in \mathbb{B}^L = \{0, 1\}^L$, the Hamming distance $d_\mathbb{H}(b_i, b_j)$ can be converted to a normalized similarity score $S_\mathbb{H}(b_i, b_j) \in [0, 1]$\text{:}
    \begin{equation}
        S_\mathbb{H}(b_i, b_j) = \frac{1}{L} (L - d_\mathbb{H}(b_i, b_j)).
    \end{equation}
    This score represents the fraction of matching bits between the two codes.
\end{definition}


Then, the problem of \textit{trajectory category hashing retrieval} we considered in this paper is defined as follows:
\begin{definition}[\textbf{Category-based trajectory hashing retrieval}]
    Given a trajectory database $\mathcal{D}$, a query trajectory $\mathcal{T}'$, and an integer $n$, the goal is to efficiently retrieve a set $\mathcal{R} \subset \mathcal{D}$ of the $n$ most similar trajectories to $\mathcal{T}'$ based on category similarity. This is achieved by:
    \begin{enumerate}
        \item Pre-compute the hash codes for all trajectories in the database as $\{b_i = \mathcal{H}(\mathcal{T}_i) | \mathcal{T}_i \in \mathcal{D}\}$.
        \item Compute the hash code for the query, $b' = \mathcal{H}(\mathcal{T}')$ and rank all trajectories $\mathcal{T}_i\in\mathcal{D}$ in ascending order of their Hamming distance to the query, $d_\mathbb{H}(b_i, b')$.
        \item Return the top-$n$ trajectories from this ranked list.
    \end{enumerate}
\end{definition}

The efficiency of this retrieval process relies on the fact that the Hamming distance computation in binary space is extremely fast, typically involving only bitwise XOR operations~\cite{wang2017survey,liu2023refining}.

\section{The Proposed Framework}

In this section, we first introduce our core insight that motivates our framework design. Next, we formalize its objective from the vector quantization technique and detail the complete algorithmic framework of GeoPTH. Finally, we analyze a key property of its binary space that guarantees the consistency of geometric similarity relations.

\subsection{Insight: Geometric Proxy for Trajectory Categories}

Our approach is built upon a core insight: trajectories that share the same category, i.e., drawn from the same underlying distribution, typically exhibit a high degree of geometric similarity. This principle, that geometric structure is a strong proxy for categories in trajectory data, is implicitly or explicitly supported by a wide range of existing research across various tasks.
For example, in trajectory-based entity linking, the unique identity of a moving object is found to be most effectively captured by its spatial signature, which is a purely geometric representation of its movement patterns~\cite{jin2020trajectory}. Similarly, the discovery of high-level category events, such as traffic jams or public gatherings, relies on identifying groups of trajectories that form dense, spatially stable clusters, with stability often measured by geometric metrics like the \textit{Hausdorff distance}~\cite{zheng2013discovery}.
Numerous state-of-the-art learning-based measures rely on grid-based representations as a foundational component~\cite{yao2019computing,yang2021t3s,yao2022trajgat,deng2024learning}. By discretizing space into cells, these methods fundamentally operate on a quantized form of geometric similarity to learn effective embeddings.

This reliance on geometric proxies is a recurring theme, suggesting that unlike in domains like computer vision, a costly, complex feature learning process to discover abstract categories may be unnecessary. Instead, a method that can directly and efficiently capture the essential geometric footprint of trajectories holds the potential for highly effective, category-based retrieval. Based on this, we introduce GeoPTH, a framework that embodies this insight by using geometric prototypes as its core hashing mechanism.

\subsection{The GeoPTH Framework: A Vector Quantization Approach}

\begin{figure}[t]
    \centering
    \includegraphics[width=0.96\columnwidth]{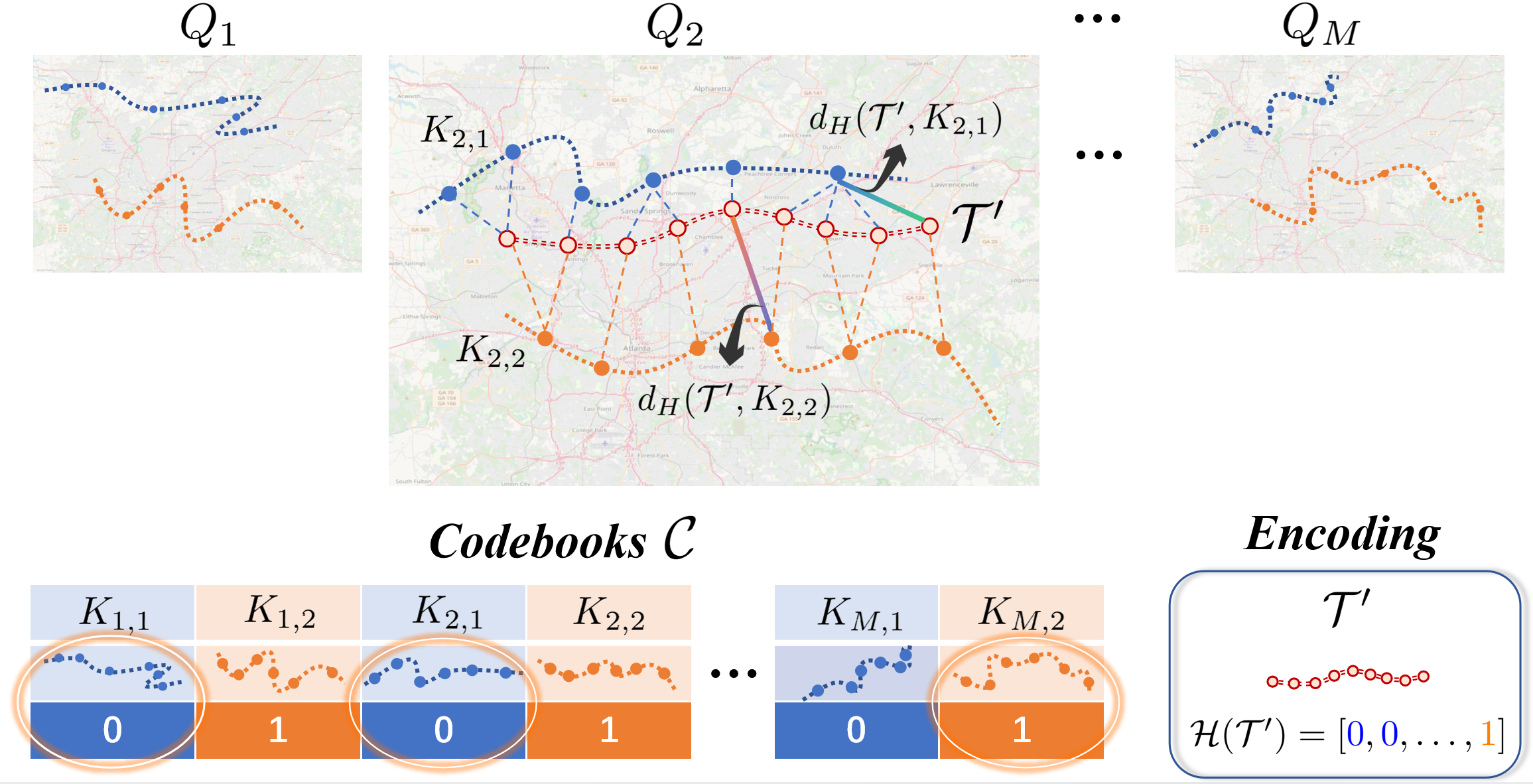}
    \caption{An illustration of the GeoPTH framework. A given query trajectory $\mathcal{T}'$ is processed by $M$ independent quantizers. For each quantizer $m$, $\mathcal{T}'$ is compared against all prototypes in its corresponding codebook $Q_m$, and assigned to the index of the prototype with the minimum Hausdorff distance ($d_H$). The resulting $M$ indices are then converted to their binary representations and concatenated to produce the final binary code $\mathcal{H}(\mathcal{T}')$.}
    \label{fig:method}
\end{figure}

Vector quantization (VQ) is a classic technique in signal processing and data analysis that represents a large set of input vectors with a smaller set of learned prototype vectors, known as a ``codebook"~\cite{gray1984vector}. The quantization process maps any given input vector to the index of its closest prototype in the codebook. Typically, iterative algorithms such as $k$-means are employed to find an optimal codebook that minimizes the average distance between the input vectors and their assigned prototypes, a quantity known as the quantization error~\cite{guo2020accelerating,lee2024brb}.

We formulate the task of trajectory hashing as a VQ problem, which partitions the unstructured space of variable-length trajectories using a codebook of prototypes. However, employing a traditional iterative optimization to find this codebook would introduce a high computational overhead. Inspired by efficient indexing structures like HVS~\cite{lu2021hvs}, which adopt direct, non-iterative partitioning of the space, we instead construct our codebooks heuristically using trajectory prototypes. This data-driven approach is both direct and highly efficient, sidestepping the need for a costly optimization process. Figure~\ref{fig:method} illustrates the GeoPTH framework, which consists of two main stages as follows:

\paragraph{Prototype Codebooks Construction}
The final $L$-bit binary codes in GeoPTH are produced by concatenating the outputs of $M$ independent, $\omega$-bit sub-hash functions, where $L = \omega \times M$. Each of these sub-hash functions, {indexed by $m \in \{1, \dots, M\}$}, is constructed through a direct, and data-driven process that consists of two steps:
\begin{enumerate}
    \item {Sampling reference trajectories.} We first randomly sample, without replacement, $\psi$ unique trajectories from the database $\mathcal{D}$. Let this set of reference trajectories be $\mathcal{R}_m = \{\mathcal{T}_{m,1}, \mathcal{T}_{m,2}, \dots, \mathcal{T}_{m,\psi}\}$.
    \item {Constructing prototype codebook.} For each sampled trajectory $\mathcal{T}_{m,j}$, we construct its corresponding prototype $K_{m,j}$ by randomly sampling $k$ points from it. The resulting set $Q_m = \{K_{m,1}, \dots, K_{m,\psi}\}$ serves as the codebook for the $m$-th quantizer. The codebook size is set to $\psi=2^{\omega}$ to ensure that the index of each prototype can be uniquely represented by a $\omega$-bit binary code, which is a strategy adopted from the "encoded hashing" mechanism in~\cite{vdeh2026xu}.
\end{enumerate}


GeoPTH constructs a prototype codebook from a set of $k$ sample points. A smaller $k$ generates a coarser prototype that enhances robustness against noise and intra-distribution variations, while a larger $k$ preserves more geometric detail, which is beneficial for distinguishing between spatially close yet structurally different distributions.
Moreover, the entire process of constructing the codebook is almost instantaneous, as its computational cost only relies on random sampling from the overall distribution of trajectories.

\paragraph{Minimum Distance Quantization}
Once the prototype codebooks $\mathcal{C}=\{Q_1, \dots, Q_M\}$ are constructed, hashing any given trajectory $\mathcal{T}$ is a deterministic quantization process. The final $L$-bit binary code $b$ is produced by concatenating the outputs of the $M$ sub-hash functions. For each sub-hash function $\mathcal{H}_m$, this is achieved by the following steps:

\begin{enumerate}
    \item {Quantization error calculation.} We compute the quantization error between the input trajectory $\mathcal{T}$ and each of the $\psi$ prototypes $K_{m,j}$ in the codebook $Q_m$ using the \textit{standard Hausdorff distance}, $d_H(\mathcal{T}, K_{m,j})$, defined as:
          \begin{equation}
              d_H(\mathcal{T}, K_{m,j}) = \max(d_h(\mathcal{T}, K_{m,j}), d_h(K_{m,j}, \mathcal{T})),
              \label{eq:dH}
          \end{equation}
          where $d_h(\cdot, \cdot)$ is the \textit{directed Hausdorff distance}, which takes the form~\cite{hausdorff}:
          \begin{equation}
              d_h(\mathcal{T}, K_{m,j}) =  \max_{p \in \mathcal{T}}\{\min_{p' \in K_{m,j}} \|p - p'\|_2 \}.
          \end{equation}

    \item {Minimum distance quantization.} We find the index $j^*$ of the prototype that minimizes this quantization error:
          \begin{equation}
              j^* = \underset{{j \in \{1, \dots, \psi\}}}{\operatorname{argmin}} d_H(\mathcal{T}, K_{m,j}).
          \end{equation}
          This step assigns the trajectory $\mathcal{T}$ to its closest prototype, which serves as a direct and practical approach to achieving the objective defined in Eq.~(\ref{eq:H*}), using geometric proximity as a proxy of category similarity. Trajectories that are geometrically similar are more likely to be mapped to the same hash index, which directly optimizes the typical objective of minimizing the expected quantization error~\cite{wu2019vector}:
          \begin{equation}
              \min_{\mathcal{H}_{m}}\mathbb{E}_{\mathcal{T}\sim\mathcal{D}}[d_H(\mathcal{T},K_{m,\mathcal{H}_{m}(\mathcal{T})})]. \label{eq:6_revised}
          \end{equation}
          It is crucial to note that while this hashing step minimizes the error for a given codebook, the codebook itself is constructed via a direct, data-driven heuristic of random sampling, rather than a costly iterative process.

    \item {Index to binary encoding.} The resulting index, $j^* - 1$, is converted into its $\omega$-bit binary representation, which becomes the sub-hash code $\mathcal{H}_m(\mathcal{T})$.
\end{enumerate}

Finally, the $M$ sub-hash codes are concatenated to form the final $L$-bit binary code $b = [\mathcal{H}_1(\mathcal{T}), \dots , \mathcal{H}_M(\mathcal{T})]$ for the trajectory $\mathcal{T}$. In essence, GeoPTH leverages the robust, shape-aware properties of the \textit{Hausdorff distance} to compare a trajectory against a set of representative geometric prototypes. This process transforms the spatial footprint of the trajectory into a compact binary code.

Conceptually, the distance function in Eq.~(\ref{eq:dH}) could be any metric that measures the dissimilarity between two point sets or trajectories, such as the \textit{DTW} and \textit{Fréchet distance}. However, our analysis reveals that the standard \textit{Hausdorff distance} is a superior choice for the GeoPTH framework. This superiority stems from its fundamental metric properties, demonstrated in the following lemma.

\begin{lemma}\cite{hausdorff}
    The \textit{Hausdorff distance} $d_H$ is a metric on the set of non-empty compact subsets of a metric space. As such, it satisfies the triangle inequality: for any three non-empty compact sets X, Y, and Z, we have $d_H(X, Z) \le d_H(X, Y) + d_H(Y, Z)$.
    \label{lem:1}
\end{lemma}

This property provides a strong theoretical guarantee for the locality-preserving nature of the quantization process. Consider any two trajectories, $\mathcal{T}$ and $\mathcal{T}'$, that are mapped to the same prototype index $j^*$ by a sub-hash function $\mathcal{H}_m$. By applying the triangle inequality from Lemma~\ref{lem:1}, we can bound the distance between these two trajectories:
\begin{equation}
    d_H(\mathcal{T}, \mathcal{T}') \le d_H(\mathcal{T}, K_{m,j^*}) + d_H(K_{m,j^*}, \mathcal{T}').
\end{equation}
This inequality demonstrates that \textit{Hausdorff distance} between any two trajectories assigned to the same prototype is bounded by the sum of their individual quantization errors with respect to that prototype. This ensures that trajectories mapped to the same hash index are indeed close in the original metric space, a fundamental requirement for effective similarity search.

Therefore, \textit{Hausdorff distance} is suitable for vector quantization used in our proposed GeoPTH framework. In contrast, \textit{DTW} does not satisfy this property, which could lead to unstable quantization errors. Although \textit{Fréchet distance} is a true metric, it is less suitable for GeoPTH. \textit{Fréchet distance} is sensitive to the sequential order of points along a trajectory. However, our prototypes, constructed by randomly sampling points, are inherently {unordered point sets} where the original sequence information is lost. This mismatch between an order-dependent metric and unordered prototypes makes the \textit{Fréchet distance} a suboptimal choice. Our experimental results in the following section demonstrate that the \textit{Hausdorff distance} yields superior performance within the GeoPTH framework compared to \textit{DTW} and \textit{Fréchet distance}.

\section{Experiment}

In this section, we compare GeoPTH with traditional metrics and state-of-the-art learning-based measures on seven real-world multi-class trajectory datasets focusing on both the effectiveness and efficiency of retrieval.

\subsection{Experimental Setup}

\begin{table}[t]
    \centering
    \resizebox{0.98\linewidth}{!}{
        \begin{tabular}{@{}lrrrr@{}}
            \toprule
            {Dataset}         & {\#Trajectories} & {\#Points}  & {min -- max $|\mathcal{T}|$} & {\#Categories} \\
            \midrule
            \texttt{Cyclists} & $265$            & $76,051$    & $52 - 1,257$                 & $3$            \\
            \texttt{Traffic}  & $300$            & $15,000$    & $50 - 50$                    & $11$           \\
            \texttt{Pedes3}   & $610$            & $202,272$   & $196 - 620$                  & $3$            \\
            \texttt{Casia}    & $1,500$          & $143,383$   & $16 - 612$                   & $15$           \\
            \texttt{Cross}    & $1,900$          & $24,420$    & $5 - 23$                     & $19$           \\
            \texttt{Gowalla}  & $15,760$         & $337,766$   & $5 - 902$                    & $200$          \\
            \texttt{Geolife}  & $86,113$         & $8,179,642$ & $90 - 100$                   & $12$           \\
            \bottomrule
        \end{tabular}}
    \caption{Statistics of the real-world trajectory datasets used. $|\mathcal{T}|$ denotes the length of a trajectory.}
    \label{tab:datasets}
\end{table}


\subsubsection{Datasets}
Our experimental evaluation is conducted on seven established multi-class benchmark datasets for trajectory retrieval: \texttt{Casia}~\cite{casia_dataset}, \texttt{Cross}~\cite{cross_dataset}, \texttt{Cyclists}, \texttt{Pedes3}~\cite{wang2023tidkc}, \texttt{Traffic}~\cite{traffic_dataset}, \texttt{Geolife}~\cite{zheng2010geolife}, and \texttt{Gowalla}~\cite{Gowalla}. The first five datasets contain trajectory categories corresponding to different behaviors or entities and have been previously employed in studies on anomaly detection and clustering. The \texttt{Geolife} dataset comprises trajectories recorded from 182 users in Beijing, China, over a five-year period. Its category labels are derived from the methodology in E$^2$DTC~\cite{fang2021e2dtc}, a method that classifies trajectories by grouping those that are spatially close.
The \texttt{Gowalla} dataset contains check-in sequences from 196,591 users over one year. In this dataset, each user's activity is typically concentrated within a specific geographic area, and these areas are well-separated from one another. We select the trajectories from the 200 most active users, then discard any of these trajectories with a length of less than five points. Following the protocol of~\cite{Gowalla}, each user corresponds to a unique category, which represents its daily check-in behavior trajectory. The statistical details of these benchmark datasets are summarized in Table~\ref{tab:datasets}.

To configure our experiments, we employ a data partitioning strategy that varies based on the size of the dataset. For the datasets containing fewer than 10,000 trajectories, we designate 25\% of the data as the query set and the remaining 75\% as the candidate database. To ensure effective training for learning-based approaches, this 75\% database partition is used as the training set. Following \textit{GnesDA}~\cite{chang2025gnesda},
for the two larger datasets with more than 10,000 trajectories, we adopt the same partitioning methodology that randomly sampling 1,000 trajectories to form the query set and 10,000 for the database. A dedicated training set of 3,000 trajectories is then sampled from the remaining data.




\begin{table}[t]
    \centering
    \resizebox{0.38\textwidth}{!}{
        \begin{tabular}{@{}l cccc @{}}
            \toprule
            Dataset           & GeoPTH                         & Hausdorff         & Fr\'echet         & DTW              \\
            \midrule
            \texttt{Cyclists} & \textbf{0.971}$_{\pm .018}$
                              & \underline{0.929}              & $0.867$           & $0.825$                              \\
            \texttt{Traffic}  & $0.964_{\pm .012}$
                              & \underline{0.979}              & \textbf{0.992}    & $0.949$                              \\
            \texttt{Pedes3}   & \textbf{0.929}$_{\pm .003}$
                              & \underline{0.917}              & $0.832$           & $0.845$                              \\
            \texttt{Casia}    & \underline{0.909}$_{\pm .003}$
                              & $0.804$                        & $0.808$           & \textbf{0.918}                       \\
            \texttt{Cross}    & \underline{0.959}$_{\pm .002}$
                              & $0.954$                        & $0.956$           & \textbf{0.967}                       \\
            \texttt{Gowalla}  & $0.277_{\pm .002}$             & \underline{0.451} & $0.431$           & \textbf{0.459}   \\
            \texttt{Geolife}  & \textbf{0.975}$_{\pm .003}$
                              & $0.914$                        & $0.910$           & \underline{0.971}                    \\
            \midrule
            Avg. Rank         & \textbf{2.00}                  & 2.71              & 3.00              & \underline{2.29} \\
            \bottomrule
        \end{tabular}}
    \caption{Retrieval performance comparison in terms of mAP between GeoPTH and traditional baseline metrics. The results for GeoPTH correspond to the code length of $L=64$, shown as mean $\pm$ 2*SE (Standard Error). For each row, the best result is in \textbf{bold}, and the second-best is \underline{underlined}.}
    \label{tab:tra}
\end{table}


\begin{table*}[t]
    \centering
    \resizebox{\textwidth}{!}{
        \begin{tabular}{@{}c l c cc cc cc cc cc@{}}
            \toprule
                                          &                   & \multirow{2}{*}{GeoPTH}        & \multicolumn{2}{c}{Traj2Hash}  & \multicolumn{2}{c}{GnesDA} & \multicolumn{2}{c}{TrajGAT} & \multicolumn{2}{c}{NeuTraj} & \multicolumn{2}{c}{t2vec}                                                                                                                                                   \\
            \cmidrule(lr){4-5} \cmidrule(lr){6-7} \cmidrule(lr){8-9} \cmidrule(lr){10-11} \cmidrule(lr){12-13}
                                          & Dataset           &                                & dense                          & binary                     & dense                       & binary                      & dense                          & binary                         & dense                          & binary             & dense                          & binary             \\
            \midrule
            \multirow{7}{*}{\rotatebox{90}{\textbf{L/d=32}}}
                                          & \texttt{Cyclists} & \underline{0.962}$_{\pm .015}$ & $0.883_{\pm .007}$             & $0.805_{\pm .006}$         & $0.843_{\pm .015}$          & $0.757_{\pm .020}$          & \textbf{0.989}$^*_{\pm .001}$  & $0.911_{\pm .006}$             & $0.760_{\pm .031}$             & $0.758_{\pm .033}$ & $0.673_{\pm .007}$             & $0.678_{\pm .018}$ \\
                                          & \texttt{Traffic}  & $0.956_{\pm .021}$             & \underline{0.961}$_{\pm .008}$ & $0.854_{\pm .005}$         & \textbf{0.969}$_{\pm .003}$ & $0.740_{\pm .016}$          & $0.889_{\pm .021}$             & $0.756_{\pm .015}$             & $0.812_{\pm.042}$              & $0.730_{\pm .037}$ & $0.728_{\pm .036}$             & $0.420_{\pm .050}$ \\
                                          & \texttt{Pedes3}   & $0.925_{\pm .011}$             & $0.910_{\pm .002}$             & $0.855_{\pm .020}$         & $0.888_{\pm .003}$          & $0.856_{\pm .004}$          & \textbf{0.986}$_{\pm .001}$    & \underline{0.941}$_{\pm .009}$ & $0.695_{\pm .020}$             & $0.694_{\pm .021}$ & $0.674_{\pm .033}$             & $0.630_{\pm .029}$ \\
                                          & \texttt{Casia}    & \textbf{0.865}$_{\pm .011}$    & $0.587_{\pm .037}$             & $0.525_{\pm .048}$         & $0.526_{\pm .012}$          & $0.402_{\pm .022}$          & \underline{0.861}$_{\pm .015}$ & $0.739_{\pm .028}$             & $0.602_{\pm .011}$             & $0.597_{\pm .014}$ & $0.614_{\pm .011}$             & $0.605_{\pm .012}$ \\
                                          & \texttt{Cross}    & \textbf{0.934}$_{\pm .009}$    & $0.805_{\pm .017}$             & $0.574_{\pm .033}$         & $0.791_{\pm .004}$          & $0.763_{\pm .016}$          & \underline{0.898}$_{\pm .018}$ & $0.813_{\pm .026}$             & $0.593_{\pm .037}$             & $0.566_{\pm .037}$ & $0.625_{\pm .018}$             & $0.591_{\pm .009}$ \\
                                          & \texttt{Gowalla}  & $0.219_{\pm .006}$             & $0.124_{\pm .006}$             & $0.046_{\pm .002}$         & $0.162_{\pm .004}$          & $0.052_{\pm .006}$          & \textbf{0.304}$_{\pm .007}$    & $0.167_{\pm .018}$             & $0.066_{\pm .004}$             & $0.066_{\pm .002}$ & \underline{0.252}$_{\pm .002}$ & $0.231_{\pm .005}$ \\
                                          & \texttt{Geolife}  & \textbf{0.962}$_{\pm .015}$    & $0.401_{\pm .011}$             & $0.330_{\pm .012}$         & $0.231_{\pm .069}$          & $0.190_{\pm .014}$          & $0.541_{\pm .078}$             & $0.347_{\pm .034}$             & \underline{0.640}$_{\pm .062}$ & $0.527_{\pm .103}$ & $0.604_{\pm .009}$             & $0.530_{\pm .009}$ \\
            \midrule
            \multirow{7}{*}{\rotatebox{90}{\textbf{L/d=64}}}
                                          & \texttt{Cyclists} & \textbf{0.971}$_{\pm .018}$    & $0.913_{\pm .003}$             & $0.902_{\pm .006}$         & $0.807_{\pm .018}$          & $0.759_{\pm .018}$          & \underline{0.965}$_{\pm .016}$ & $0.908_{\pm .039}$             & $0.779_{\pm .026}$             & $0.776_{\pm .021}$ & $0.641_{\pm .027}$             & $0.645_{\pm .040}$ \\
                                          & \texttt{Traffic}  & \textbf{0.964}$_{\pm .012}$    & $0.962_{\pm .005}$             & $0.894_{\pm .006}$         & \textbf{0.964}$_{\pm.002}$  & $0.890_{\pm .016}$          & $0.912_{\pm .032}$             & $0.820_{\pm .026}$             & $0.866_{\pm .006}$             & $0.775_{\pm .026}$ & $0.707_{\pm .012}$             & $0.427_{\pm .021}$ \\
                                          & \texttt{Pedes3}   & $0.929_{\pm .003}$             & $0.910_{\pm .002}$             & $0.837_{\pm .012}$         & $0.887_{\pm .002}$          & $0.854_{\pm .008}$          & \textbf{0.988}$^*_{\pm .005}$  & \underline{0.940}$_{\pm.010}$  & $0.642_{\pm .035}$             & $0.655_{\pm.031}$  & $0.681_{\pm .011}$             & $0.671_{\pm .022}$ \\
                                          & \texttt{Casia}    & \textbf{0.909}$_{\pm .003}$    & $0.629_{\pm .035}$             & $0.567_{\pm .046}$         & $0.520_{\pm .012}$          & $0.384_{\pm .027}$          & \underline{0.838}$_{\pm .016}$ & $0.751_{\pm .021}$             & $0.600_{\pm .011}$             & $0.610_{\pm .011}$ & $0.658_{\pm .004}$             & $0.660_{\pm .010}$ \\
                                          & \texttt{Cross}    & \textbf{0.959}$_{\pm .002}$    & $0.837_{\pm .002}$             & $0.698_{\pm .018}$         & $0.780_{\pm .005}$          & $0.745_{\pm .008}$          & \underline{0.903}$_{\pm .015}$ & $0.829_{\pm .026}$             & $0.736_{\pm .035}$             & $0.723_{\pm .011}$ & $0.591_{\pm .023}$             & $0.590_{\pm .005}$ \\
                                          & \texttt{Gowalla}  & \underline{0.277}$_{\pm .002}$ & $0.174_{\pm .019}$             & $0.071_{\pm .011}$         & $0.148_{\pm .009}$          & $0.043_{\pm .003}$          & \textbf{0.298}$_{\pm .026}$    & $0.179_{\pm .027}$             & $0.079_{\pm .004}$             & $0.086_{\pm.005}$  & $0.263_{\pm .003}$             & $0.258_{\pm .005}$ \\
                                          & \texttt{Geolife}  & \textbf{0.975}$_{\pm .003}$    & $0.453_{\pm .010}$             & $0.411_{\pm .008}$         & $0.168_{\pm .011}$          & $0.152_{\pm .023}$          & $0.536_{\pm .122}$             & $0.371_{\pm .096}$             & \underline{0.720}$_{\pm .011}$ & $0.467_{\pm .042}$ & $0.611_{\pm .023}$             & $0.578_{\pm .012}$ \\
            \midrule
            \multirow{7}{*}{\rotatebox{90}{\textbf{L/d=128}}}
                                          & \texttt{Cyclists} & \textbf{0.972}$_{\pm .012}$    & $0.913_{\pm .113}$             & $0.894_{\pm .019}$         & $0.793_{\pm .014}$          & $0.708_{\pm .026}$          & \textbf{0.972}$_{\pm .016}$    & $0.878_{\pm .010}$             & $0.770_{\pm .049}$             & $0.768_{\pm.029}$  & $0.605_{\pm .026}$             & $0.591_{\pm .035}$ \\
                                          & \texttt{Traffic}  & \textbf{0.982}$^*_{\pm .004}$  & \underline{0.962}$_{\pm .009}$ & $0.923_{\pm .007}$         & $0.928_{\pm.022}$           & $0.759_{\pm .042}$          & $0.830_{\pm .036}$             & $0.799_{\pm .047}$             & $0.881_{\pm .019}$             & $0.692_{\pm .012}$ & $0.686_{\pm .026}$             & $0.410_{\pm .029}$ \\
                                          & \texttt{Pedes3}   & \underline{0.941}$_{\pm .004}$ & $0.898_{\pm .005}$             & $0.859_{\pm .005}$         & $0.881_{\pm .003}$          & $0.841_{\pm .017}$          & \textbf{0.984}$_{\pm .003}$    & $0.933_{\pm .022}$             & $0.622_{\pm .040}$             & $0.616_{\pm.038}$  & $0.678_{\pm .004}$             & $0.668_{\pm .003}$ \\
                                          & \texttt{Casia}    & \textbf{0.927}$^*_{\pm .002}$  & $0.690_{\pm .020}$             & $0.646_{\pm .024}$         & $0.498_{\pm .010}$          & $0.234_{\pm .020}$          & \underline{0.772}$_{\pm .027}$ & $0.714_{\pm .030}$             & $0.624_{\pm .007}$             & $0.620_{\pm .014}$ & $0.686_{\pm .012}$             & $0.684_{\pm .014}$ \\
                                          & \texttt{Cross}    & \textbf{0.979}$^*_{\pm .002}$  & $0.810_{\pm .028}$             & $0.724_{\pm .030}$         & $0.723_{\pm .021}$          & $0.536_{\pm .072}$          & \underline{0.877}$_{\pm .012}$ & $0.826_{\pm .017}$             & $0.754_{\pm .024}$             & $0.819_{\pm .007}$ & $0.615_{\pm .026}$             & $0.616_{\pm .016}$ \\
                                          & \texttt{Gowalla}  & \textbf{0.348}$^*_{\pm .004}$  & $0.158_{\pm .001}$             & $0.089_{\pm .008}$         & $0.148_{\pm .006}$          & $0.031_{\pm .003}$          & $0.285_{\pm .007}$             & $0.149_{\pm .012}$             & $0.097_{\pm .004}$             & $0.080_{\pm .004}$ & \underline{0.303}$_{\pm .002}$ & $0.294_{\pm .003}$ \\
                                          & \texttt{Geolife}  & \textbf{0.987}$^*_{\pm .001}$  & $0.503_{\pm .024}$             & $0.492_{\pm .027}$         & $0.142_{\pm .001}$          & $0.137_{\pm .016}$          & $0.559_{\pm .098}$             & $0.417_{\pm .077}$             & \underline{0.863}$_{\pm .022}$ & $0.443_{\pm .031}$ & $0.665_{\pm .009}$             & $0.671_{\pm .010}$ \\
            \midrule
            \multicolumn{2}{c}{Avg. Rank} & \textbf{1.57}     & $4.62$                         & $7.38$                         & $6.38$                     & $9.00$                      & \underline{2.52}            & $4.86$                         & $6.76$                         & $8.24$                         & $6.95$             & $7.57$                                              \\
            \bottomrule
        \end{tabular}%
    }
    \caption{Performance comparison in terms of mAP $\pm$ 2*SE against state-of-the-art learning-based measures. The comparison is evaluated across various binary code lengths ($L$) and embedding dimensions $d$, i.e., $L/d \in \{32, 64, 128\}$ \textbf{(denoting settings where $L=d$)}. We evaluate both the original ``dense" embeddings and their ``binary" counterparts. For each row, the best result is in \textbf{bold}, and the second-best is \underline{underlined}. * The overall best result for each dataset across all settings.}
    \label{tab:map_by_dimension}
\end{table*}

\subsubsection{Baselines}
To evaluate our proposed method, GeoPTH, we compare it against two categories of baselines: traditional metrics and learning-based measures. The first category comprises three traditional metrics: \textit{Hausdorff distance}, \textit{Fr\'echet distance}, and \textit{DTW}. These metrics are widely-recognized, non-learning standards for directly measuring similarity between raw trajectory data. The second category consists of state-of-the-art learning-based methods, including \textit{Traj2Hash}~\cite{deng2024learning}, \textit{GnesDA}~\cite{chang2025general}, \textit{TrajGAT}~\cite{yao2022trajgat}, \textit{NeuTraj}~\cite{yao2019computing}, and \textit{t2vec}~\cite{li2018t2vec}.
These learning-based methods accelerate the retrieval process by learning fixed-length vector representations, i.e., embeddings, that approximate traditional metrics.
To achieve this, all of these methods, with the exception of \textit{t2vec}, require a pre-computed ground-truth distance matrix (e.g. \textit{DTW}) to serve as training supervision.
Among them, \textit{Traj2Hash} is unique as it adds a layer on top of its vector output to explicitly map the vectors into binary codes via a binarization function.

To ensure a fair comparison in Hamming space, we follow the procedure from \textit{Traj2Hash}~\cite{deng2024learning} and apply its binarization process to the vector embeddings of all other learning-based methods for evaluation. In addition, we strictly follow the parameter search and settings of all learning-based methods and obtain embeddings of different dimensions (32, 64, and 128) by varying the output layer size. For GeoPTH, the number of sample trajectories $\psi$ was searched in $\{2^1,2^2,\dots,2^6\}$, and the number of points in a prototype $k$ was searched in $\{1,5,10,15,20\}$. All experiments are conducted on the same Linux machine: AMD 128-core CPU with each core running at 2 GHz and 1 TB RAM and one RTX A6000 GPU with 48GB VRAM.

\begin{table*}[t]
    \centering
    \resizebox{0.78\textwidth}{!}{
        \begin{tabular}{@{}l r rrrrrrrrr @{}}
            \toprule
                              & \multicolumn{4}{c}{CPU} & \multicolumn{5}{c}{GPU}                                                                                      \\
            \cmidrule(lr){2-5} \cmidrule(lr){6-10}
            Dataset           & GeoPTH                  & Hausdorff               & Fr\'echet & DTW             & Traj2Hash      & GnesDA & TrajGAT & NeuTraj & t2vec  \\
            \midrule
            \texttt{cyclists} & $\textbf{2}$            & \underline{6}           & $12$      & $6$             & $29$           & $47$   & $64$    & $214$   & $96$   \\
            \texttt{TRAFFIC}  & \textbf{2}              & \underline{5}           & $10$      & \underline{5}   & $11$           & $48$   & $98$    & $23$    & $171$  \\
            \texttt{pedes3}   & \textbf{3}              & \underline{22}          & $50$      & $24$            & $51$           & $106$  & $416$   & $442$   & $236$  \\
            \texttt{CASIA}    & \textbf{3}              & \underline{69}          & $145$     & $70$            & $87$           & $238$  & $239$   & $611$   & $374$  \\
            \texttt{cross}    & \textbf{5}              & $107$                   & $103$     & $108$           & \underline{42} & $269$  & $208$   & $76$    & $465$  \\
            \texttt{Gowalla}  & \textbf{8}              & $446$                   & $681$     & \underline{435} & $504$          & $624$  & $562$   & $813$   & $627$  \\
            \texttt{Geolife}  & \textbf{9}              & $469$                   & $746$     & \underline{447} & $611$          & $619$  & $556$   & $608$   & $927$  \\
            \midrule
            Avg. Rank         & \textbf{1.00}           & \underline{2.71}        & $5.29$    & $2.86$          & $4.43$         & $6.57$ & $6.57$  & $7.14$  & $8.14$ \\
            \bottomrule
        \end{tabular}
    }
    \caption{Efficiency comparison in terms of average execution time (in seconds) on all datasets. For GeoPTH and all learning-based measures, the reported times correspond to the setting with $L/d=64$. }
    \label{tab:time_comparison_by_dimension}
\end{table*}

\subsubsection{Evaluation Protocol}
We use Mean Average Precision (mAP), a standard metric for evaluating performance in category-based retrieval tasks, as the measure of accuracy~\cite{wang2017survey,luo2023survey}. The mAP is the mean of the Average Precision (AP) scores over the set of all queries $\mathcal{D}$. The AP for a single query is defined as:
$$
    AP = \frac{\sum_{n=1}^{N} (P(n) \times \text{rel}(n))}{\text{Number of relevant items}},
$$
where $N$ is the number of trajectories in the database, $P(n)$ is the precision at cut-off $n$, and $\text{rel}(n)$ is an indicator function that is 1 if the item at rank $n$ is relevant and 0 otherwise. The final mAP is then calculated as:
$$
    mAP = \frac{1}{|\mathcal{D}|} \sum_{\mathcal{T} \in \mathcal{D}} AP(\mathcal{T}).
$$

\subsection{Experimental Results}

\subsubsection{Comparison with Traditional Baseline Metrics}

We first compare the retrieval performance of GeoPTH against the traditional metrics. As shown in Table~\ref{tab:tra}, GeoPTH's performance is highly competitive, achieving the best results on three datasets (\texttt{Cyclists}, \texttt{Pedes3}, \texttt{Geolife}).
On \texttt{Gowalla}, GeoPTH performs worse due to the complex data distribution, which requires a larger code length ($L=256$) to achieve comparable results.
A particularly noteworthy observation is that GeoPTH achieves better performance than \textit{Hausdorff distance} in five out of seven benchmarks, despite using it in its core quantization function.
This is mainly because GeoPTH constructs prototypes from sampled points in trajectories and obtains the final codebooks by ensembling multiple sets of sampled trajectories, which makes it insensitive to local outliers and noise points. In contrast, it is a well-known issue that the \textit{Hausdorff distance} is sensitive to outliers and noise points, as its value is determined by the single point with the maximum deviation~\cite{taha2015efficient}.
This observation is consistent with our parameter sensitivity analysis, which shows that the model is highly robust to the choice of $k$ and achieves strong performance even with a small prototype size.

\subsubsection{Comparison with Learning-based Measures}

We then compare the retrieval performance of GeoPTH against the learning-based methods, shown in Table~\ref{tab:map_by_dimension}. The results show that GeoPTH achieves the best mAP results in the most cases. Among all seven datasets, GeoPTH achieved the best values on five datasets, while \textit{TrajGAT} achieved the best values on the remaining two datasets.
A key observation from Table~\ref{tab:map_by_dimension} is the performance degradation that occurs when converting the ``dense" embeddings of all learning-based methods into ``binary" codes including \textit{TrajGAT}. For nearly all models, this binarization process results in a substantial drop in retrieval accuracy, which is often considered a trade-off to gain the efficiency of Hamming space retrieval~\cite{deng2024learning}.
Furthermore, as these learning-based methods are designed to approximate existing traditional metrics, their results are generally worse than those of traditional metrics in Table~\ref{tab:tra}.
In contrast, as a quantization-native framework, GeoPTH generates binary codes as a direct result of its mapping process, which allows it to avoid the performance degradation.
As a quantization-based method, GeoPTH exhibits a desirable scaling property: its retrieval accuracy consistently improves as the hash code length $L$ increases from 32 to 128, allowing for a finer-grained representation of the trajectory space.

\subsubsection{Efficiency Comparison}

Table~\ref{tab:time_comparison_by_dimension} shows the efficiency comparison in terms of average query execution time, which includes all overheads such as training and mapping processes for GeoPTH and learning-based methods. The traditional metrics and GeoPTH were executed on a CPU and 16 cores are used for parallel processing, while learning-based methods were executed on a GPU. Even with GPU acceleration, many learning-based methods exhibit significant latency, sometimes failing to outperform the much simpler CPU-based traditional metrics. In contrast, GeoPTH is overwhelmingly superior, dominating all competitors across all datasets. The performance gap is particularly pronounced on large-scale datasets like \texttt{Geolife} and \texttt{Gowalla}, where GeoPTH is faster by one to two orders of magnitude, demonstrating its exceptional efficiency.

\begin{figure}[t]
    \centering
    \begin{subfigure}{0.23\textwidth}
        \centering
        \includegraphics[width=\linewidth]{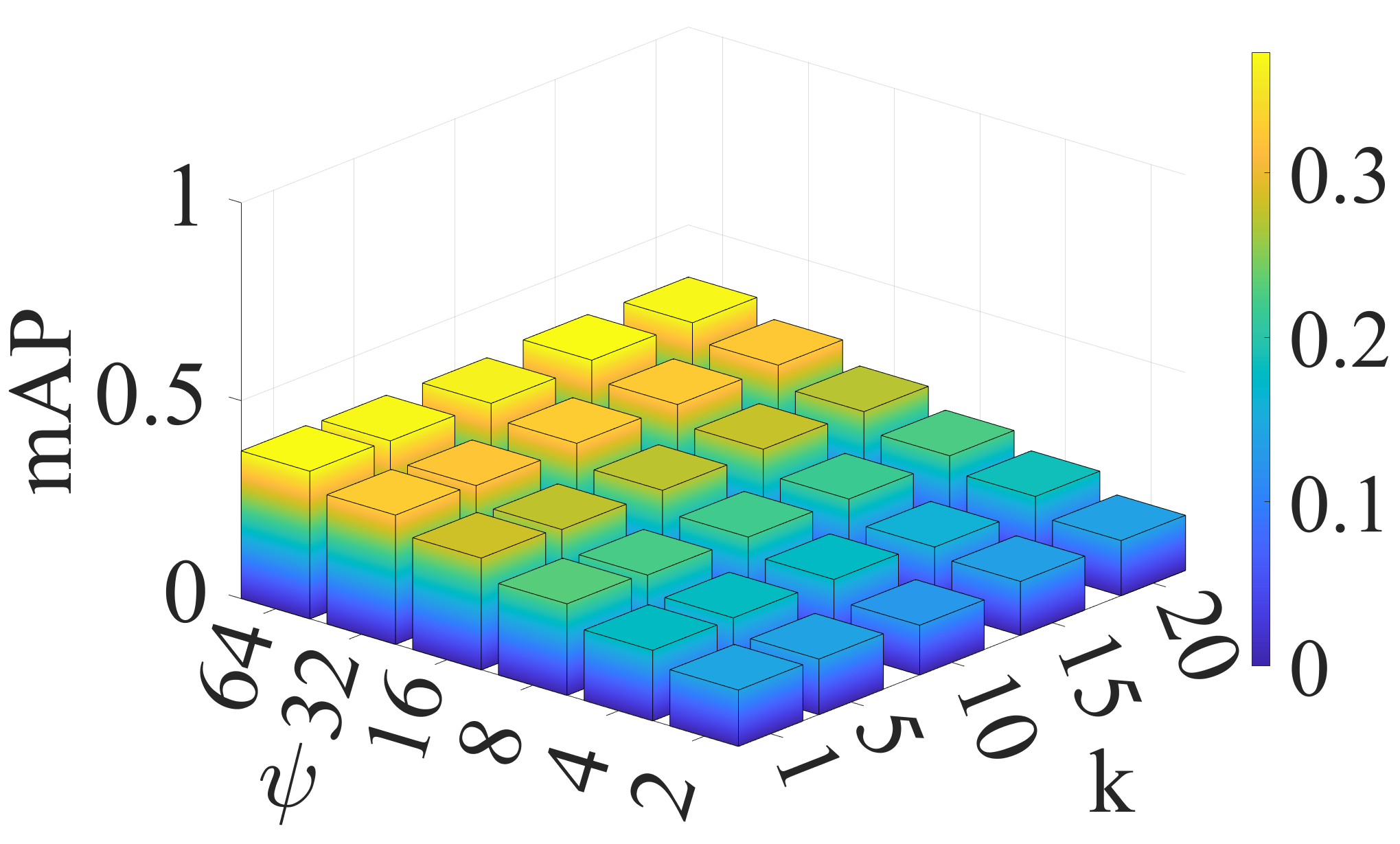}
        \caption{\texttt{Gowalla}}
        \label{fig:Gowalla}
    \end{subfigure}
    \begin{subfigure}{0.23\textwidth}
        \centering
        \includegraphics[width=\linewidth]{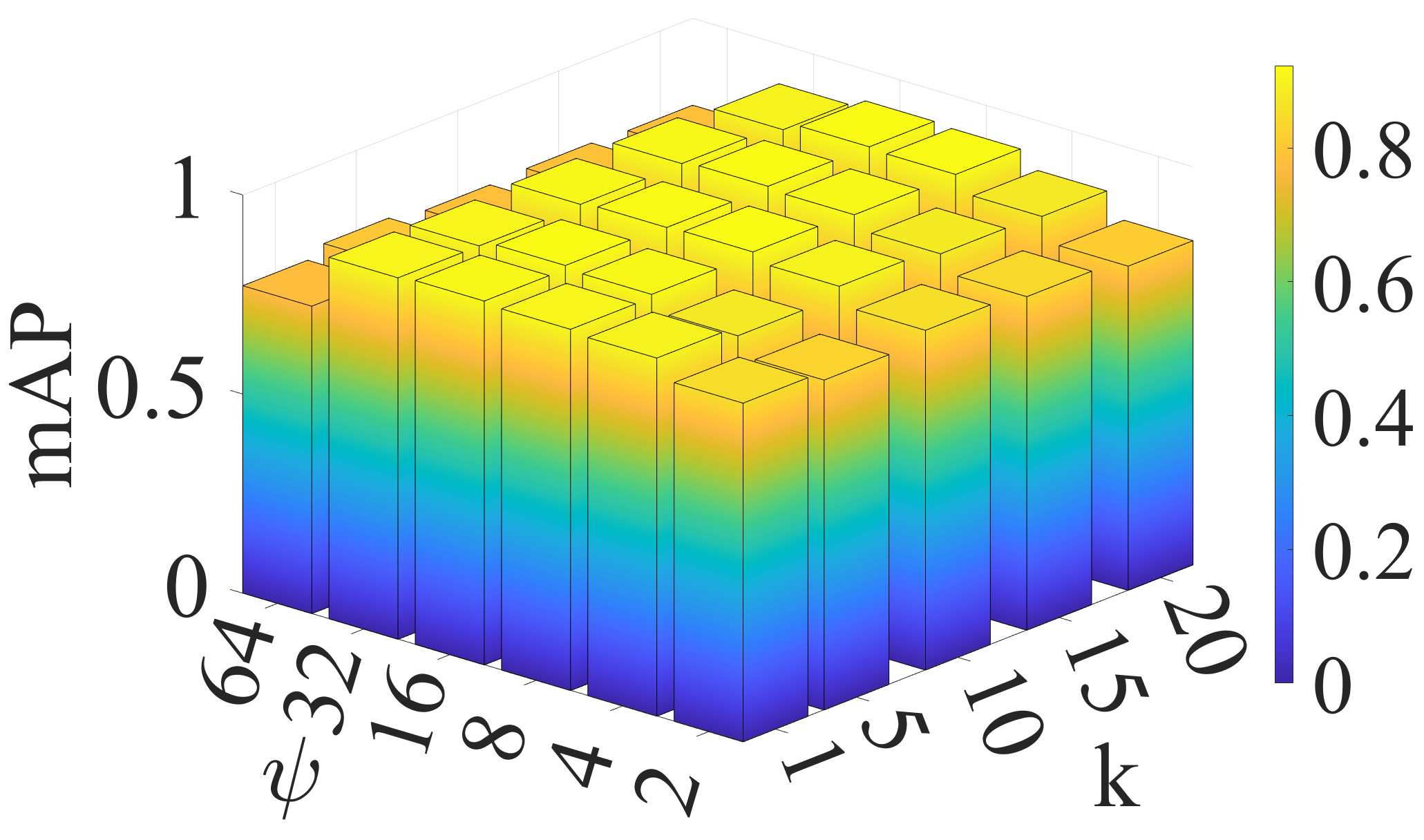}
        \caption{\texttt{Geolife}}
        \label{fig:Geolife}
    \end{subfigure}
    \caption{Parameter analysis of GeoPTH on the \texttt{Gowalla} and \texttt{Geolife} dataset. The results correspond to the code length of $L=64$.}
    \label{fig:parameter analysis}
\end{figure}

\subsection{Parameter Analysis}
We analyze the effect of GeoPTH's two key parameters on the \texttt{Gowalla} and \texttt{Geolife} datasets, which contain the largest number of trajectories in our benchmark: the codebook size $\psi$, varied across $\{2^1,2^2,\dots,2^6\}$, and the prototype size $k$, tested with values from $\{1, 5, 10, 15, 20\}$. These experiments are run with a fixed hash code length of $L=64$, and each parameter setting is evaluated 10 times to obtain an average mAP, as visualized in Figure~\ref{fig:parameter analysis}. The results suggest that while a larger codebook size $\psi$ is generally beneficial for accuracy, increasing it beyond an optimal point may lead to a slight decline in performance, as observed in Figure~\ref{fig:Geolife}. As for the prototype size $k$, the results across both datasets in Figure~\ref{fig:parameter analysis} demonstrate that the model's performance is highly robust to this parameter, achieving strong results even with a small value. Based on this analysis, we recommend setting $k=10$ for practical implementation.

\begin{table}[t]
    \centering
    \begin{tabular}{@{}l ccc @{}}
        \toprule
        Dataset           & Hausdorff                      & Fr\'echet                      & DTW                            \\
        \midrule
        \texttt{Cyclists} & \textbf{0.971}$_{\pm .018}$    & \underline{0.878}$_{\pm .004}$ & $0.839_{\pm .009}$             \\
        \texttt{Traffic}  & \underline{0.964}$_{\pm .012}$ & \textbf{0.989}$_{\pm .002}$    & $0.919_{\pm .012}$             \\
        \texttt{Pedes3}   & \textbf{0.929}$_{\pm .003}$
                          & $0.841_{\pm .009}$             & \underline{0.912}$_{\pm .002}$                                  \\
        \texttt{Casia}    & \textbf{0.909}$_{\pm .003}$    & $0.767_{\pm .005}$             & \underline{0.826}$_{\pm .010}$ \\
        \texttt{Cross}    & \textbf{0.959}$_{\pm .002}$    & $0.897_{\pm .017}$             & \underline{0.907}$_{\pm .007}$ \\
        \texttt{Gowalla}  & \textbf{0.277}$_{\pm .002}$    & \underline{0.267}$_{\pm .015}$ & $0.265_{\pm .002}$             \\
        \texttt{Geolife}  & \textbf{0.975}$_{\pm .003}$
                          & $0.925_{\pm .002}$             & \underline{0.973}$_{\pm .004}$                                  \\
        \midrule
        Avg. Rank         & \textbf{1.14}                  & \underline{2.43}               & \underline{2.43}               \\
        \bottomrule
    \end{tabular}
    \caption{Ablation study: retrieval performance of GeoPTH with different quantization metrics. The results correspond to the code length of $L=64$.}
    \label{tab:ab}
\end{table}

\subsection{Ablation Study}

To validate our choice of the core distance metric, we conduct an ablation study with results shown in Table~\ref{tab:ab}. In this experiment, we replace the \textit{Hausdorff distance} in our quantization step (Eq. 5) with \textit{Fr\'echet distance} and \textit{DTW}. The results empirically confirm our preliminary analysis, showing a consistent and significant performance degradation when using either alternative metric. This is because \textit{DTW} does not satisfy the triangle inequality crucial for stable quantization, while the order-sensitive \textit{Fr\'echet distance} is fundamentally incompatible with our unordered, sampled prototypes. For example, a trajectory that moves back and forth would be poorly measured against its simple spatial footprint by \textit{Fr\'echet distance}. This study thus validates that the \textit{Hausdorff distance} is the most theoretically sound and empirically effective choice for the GeoPTH framework.

\section{Conclusion}

In this paper, we addressed the persistent trade-off between accuracy and efficiency in trajectory similarity retrieval. We introduced GeoPTH, a novel, lightweight, and non-learning framework that generates robust binary codes for trajectories by quantizing them against geometric prototypes. Our core insight is that for trajectory data, a direct, geometry-aware hashing mechanism can bypass the need for costly and complex deep learning models. Extensive experiments demonstrated that GeoPTH not only achieves retrieval accuracy competitive with both traditional metrics and state-of-the-art learning-based methods but also delivers an overwhelming efficiency advantage, outperforming all competitors by orders of magnitude. Our work shows that a simple, prototype-centric approach offers a powerful and practical alternative, striking an exceptional balance between retrieval performance and computational cost. For future work, exploring more sophisticated prototype selection strategies and extending the framework to incorporate temporal or semantic information are promising directions.

\newpage


\bibliography{main}
\newpage

\appendix
\section{Hyperparameter Search Space}
\label{appendix:hyperparams}
In this section we list the search ranges of hyperparameters for GeoPTH and all the learning-based methods we compared, as shown in Table~\ref{tab:hyperparams}. Specifically, for GnesDA, we adopted 3 as the number of convolution layers, which is the consistently optimal value reported in the paper~\cite{chang2025gnesda}. As for t2vec, since the optimal cell size varies considerably across datasets (even exceeding the reported range in~\cite{li2018t2vec}), we performed a two-stage search. Following the initial range search outlined in Table~\ref{tab:hyperparams}, we narrowed the scope for a second round, for example, searching the set $\{1, 5, 10, 25, 50, 100\}$ within the range of 1 to 100.

\begin{table}[h]
    \centering
    \resizebox{0.48\textwidth}{!}{
        \begin{tabular}{l|l|c}
            \toprule
            Method & Parameters                         & Search Range                            \\
            \midrule
            \multirow{2}{*}{GeoPTH}
                   & the codebook size $\psi$           & $\psi\in \{ 2^1, 2^2, \cdots, 2^6 \}$   \\
                   & the prototype size $k$             & $k \in \{1, 5, 10, 15, 20\} $           \\
            \midrule
            \multirow{2}{*}{Traj2Hash}
                   & the margin $\alpha$                & $\alpha \in \{1, 5, 10\}$               \\
                   & the balanced weight $\gamma$       & $\gamma \in \{1, 5, 10\}$               \\
            \midrule
            \multirow{1}{*}{GnesDA}
                   & the convolution layers             & fixed to $3$                            \\
            \midrule
            \multirow{2}{*}{TrajGAT}
                   & the PR-quadtree threshold $\delta$ & $\delta \in \{10, 50, 100, 500, 1000\}$ \\
                   & the hierarchy layers $\eta$        & $\eta \in \{1, 2, 3\}$                  \\
            \midrule
            \multirow{1}{*}{NeuTraj}
                   & the scan width $w$                 & $w \in \{0, 1, 2, 3, 4\}$               \\
            \midrule
            \multirow{2}{*}{t2vec}
                   & the cell size                      & $\{10^{-3}, 10^{-2}, \cdots, 10^3\}$    \\
                   & the hot cell threshold $\delta$    & $\delta \in \{1, 5, 10, 25, 50\}$       \\
            \bottomrule
        \end{tabular}
    }
    \caption{Hyperparameter search ranges for GeoPTH and the learning-based methods.}
    \label{tab:hyperparams}
\end{table}

\end{document}